\documentclass{article}
\usepackage{arxiv}

\usepackage{graphicx}
\usepackage{amssymb}
\usepackage{amsmath}
\usepackage{enumerate}
\usepackage{booktabs}
\usepackage{algorithm}
\usepackage{algorithmic}
\usepackage{color}
\usepackage{colortbl}
\usepackage[colorlinks,linkcolor=red,anchorcolor=blue,citecolor=green]{hyperref}
\usepackage{bbm}
\usepackage{tabularx}
\usepackage{float}
\usepackage{subfloat}
\usepackage[caption=false,font=normalsize,labelfont=sf,textfont=sf]{subfig}
\usepackage{multirow}
\usepackage[table,xcdraw]{xcolor}
\usepackage[english]{babel}
\usepackage[sorting=none, style=ieee, citestyle=numeric-comp, natbib=true, backend=bibtex]{biblatex}
\usepackage{svg}
\usepackage{booktabs}

\newcolumntype{C}{>{\centering\arraybackslash}X}

\usepackage{authblk}

\author[1,2]{Pedram Ghamisi}
\author[1]{Weikang Yu}
\author[3,5]{Andrea~Marinoni}
\author[4]{Caroline M. Gevaert}
\author[4]{Claudio Persello}
\author[5]{Sivasakthy Selvakumaran}
\author[6]{Manuela Girotto}
\author[7]{Benjamin P. Horton}
\author[8,9]{Philippe Rufin}
\author[9]{Patrick Hostert}
\author[10]{Fabio Pacifici}
\author[2]{Peter M. Atkinson}
\affil[1]{Helmholtz-Zentrum Dresden-Rossendorf (HZDR), 09599 Freiberg, Germany} 
\affil[2]{Lancaster University, LA1 4YR Lancaster, U.K.}
\affil[3]{UiT The Arctic University of Norway, 9019 Tromsø, Norway}
\affil[4]{University of Twente, 7522 NH Enschede, Netherlands}
\affil[5]{University of Cambridge, CB3 0FA Cambridge, U.K.}
\affil[6]{University of California, Berkeley, CA 94720 Berkeley, U.S.A.}
\affil[7]{Nanyang Technological University, 639798 Singapore, Singapore}
\affil[8]{UCLouvain, 1348 Louvain-la-Neuve, Belgium}
\affil[9]{Humboldt Universität zu Berlin, 10117 Berlin, Germany}
\affil[10]{Maxar Technologies Inc, CO 80234 Westminster, U.S.A.}

\date{}    
\title{Responsible AI for Earth Observation}

\begin{document}

	%

	%
	%

	%

	\maketitle
	
	\begin{abstract}
		The convergence of artificial intelligence (AI) and Earth observation (EO) technologies has brought geoscience and remote sensing into an era of unparalleled capabilities. AI's transformative impact on data analysis, particularly derived from EO platforms, holds great promise in addressing global challenges such as environmental monitoring, disaster response and climate change analysis. However, the rapid integration of AI necessitates a careful examination of the responsible dimensions inherent in its application within these domains. In this paper, we represent a pioneering effort to systematically define the intersection of AI and EO, with a central focus on responsible AI practices. Specifically, we identify several critical components guiding this exploration from both academia and industry perspectives within the EO field: AI and EO for social good, mitigating unfair biases, AI security in EO, geo-privacy and privacy-preserving measures, as well as maintaining scientific excellence, open data, and guiding AI usage based on ethical principles. Furthermore, the paper explores potential opportunities and emerging trends, providing valuable insights for future research endeavors.
	\end{abstract}
	
	
	
	

	%
	
	\section{Introduction}
\begin{figure}
    \centering
    \includegraphics[scale=0.55]{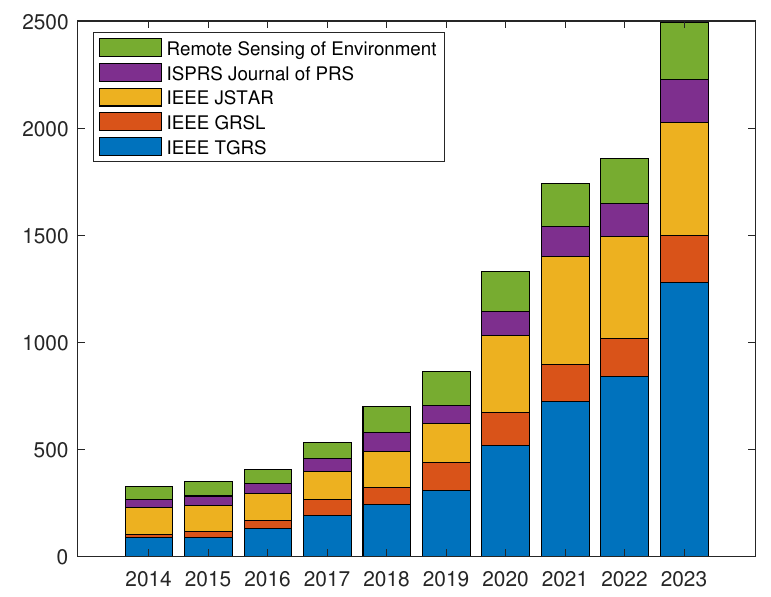}
    \caption{Number of AI4EO-related studies published in IEEE TGRS from 2014 to 2023. Data is collected from Google Scholar advanced search: keywords: ("machine learning" or "deep learning") and "remote sensing."}
    \label{fig:papers1}
\end{figure}

In the evolving landscape of EO, the fusion of AI with Earth observation (EO) holds tremendous promise for unlocking insights into our planet's dynamics \cite{ghamisi2018new}. 
Specifically, the integration of AI techniques for EO (AI4EO) significantly enhances our capacity to distill meaningful information from remotely sensed data \cite{zhu2017deep}, which became a substantial surge since 2020, as shown in Fig. \ref{fig:papers1}. 
Nonetheless, the complex nature of Earth systems requires a thorough understanding of AI's inherent limitations and potential consequences, which necessitates the implementation of responsible AI practices in EO. This shapes the particular focus of this paper. 

\begin{figure*}[!ht]
    \centering
    \includegraphics[width=\linewidth]{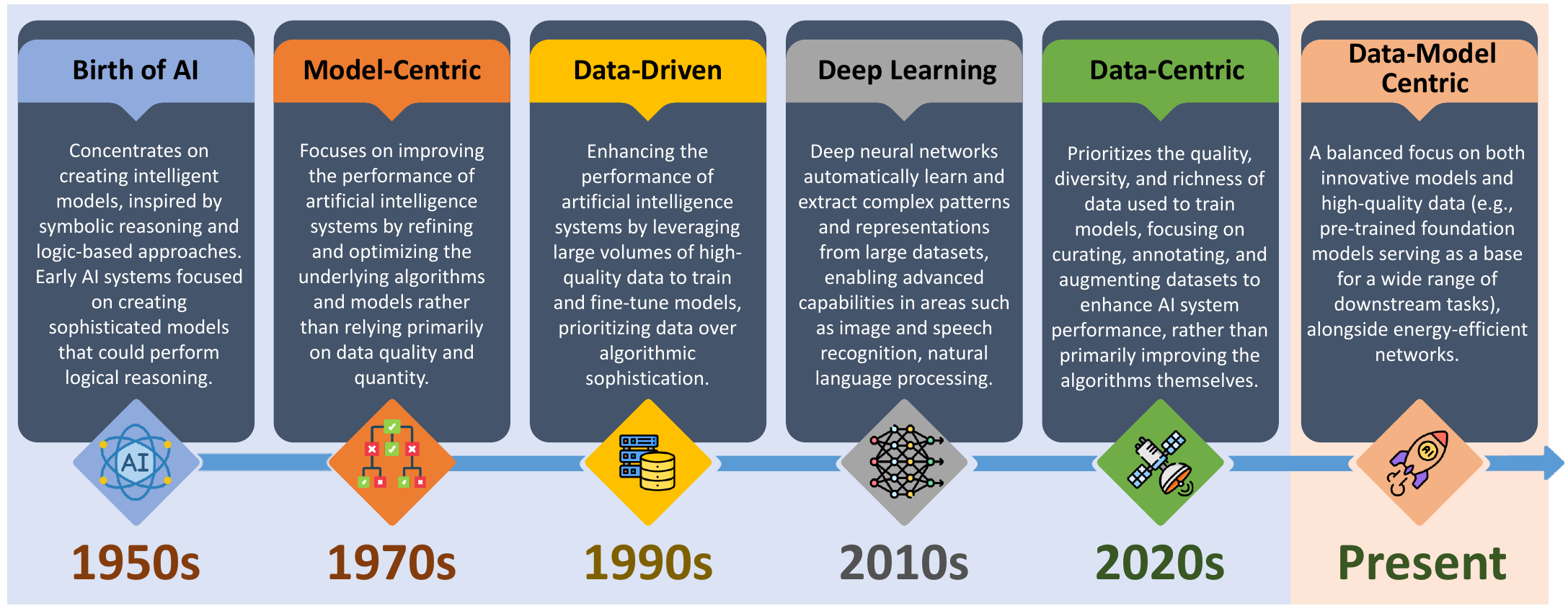}
    \caption{Overview of the development of AI algorithms over the last decades. The focus of AI algorithms has shifted multiple times between being model-centric and data-centric. The blue part illustrates the historical development of AI concerning data and models, while the orange part reflects our beliefs in the EO community regarding current and future developments. The concept of responsible AI gained significant attention in the late 2010s and early 2020s. During this period, concerns about AI biases, fairness, transparency, and accountability began to receive increased attention.}
\label{fig:intro_devAI}
\end{figure*}
Fig. \ref{fig:intro_devAI} shows an overview of the development of AI algorithms over the last decades. The birth of AI as a distinct field is often attributed to a workshop held at Dartmouth College in the summer of 1956. The concept of responsible AI gained significant attention in the late 2010s and early 2020s, and with generative-adversarial approaches as one of the catalysts. During this period, concerns about AI biases, fairness, transparency and accountability began to gain prominence, with early influential works such as "Gender Shades" (2018) \cite{buolamwini2018gender} highlighting biases in facial recognition technologies. This era also saw the release of ethical guidelines like the IEEE's "Ethically Aligned Design" (2016) \cite{shahriari2017ieee} and the ACM's updated Code of Ethics (2018) \cite{gotterbarn2018acm}. In the early 2020s, regulatory efforts intensified with the European Commission's "Ethics Guidelines for Trustworthy AI" (2019) \cite{hleg2019ethics} and the OECD's AI Principles (2019) \cite{galindo2021overview}. Major tech companies like Google and Microsoft established dedicated responsible AI teams and initiatives. The concept of responsible AI continued to evolve, integrating into AI development pipelines and emphasizing both data quality and model innovation. Standardization efforts by organizations like ISO and educational initiatives further promoted responsible AI practices, addressing ethical considerations such as fairness, transparency, and bias mitigation. In our opinion, and as indicated by the light orange color in Fig. \ref{fig:intro_devAI}, we have transitioned from an era of singular focus on either model-centric or data-centric approaches to a new era characterized by balanced data-model approaches. This paradigm emphasizes both model innovation and data equally, exemplified by the rise of foundation models \cite{mai2023opportunities, hong2023spectralgpt}. Despite this shift, we believe responsible AI continues to play an increasingly important role in guiding the development and deployment of AI technologies.


From a technical perspective, the ambition of responsible AI goes beyond theoretical discussions and becomes a vital aspect of both research and application \cite{dignum2019responsible,9527092}. Therefore, this paper articulates the use of AI4EO with a particular focus on responsible practices, providing a comprehensive guide for the ethical design and deployment of AI in geospatial and environmental contexts. Considering the centrality of human decisions in shaping AI applications, we emphasize the incorporation of domain-specific values at every decision point. As discussed in this paper, responsible AI requires a systematic process to ensure that ethical considerations are incorporated into the body of AI4EO projects, from data collection to the interpretation of geospatial intuitions.

\begin{figure}[!ht]
    \centering
    \includegraphics[scale=0.31]{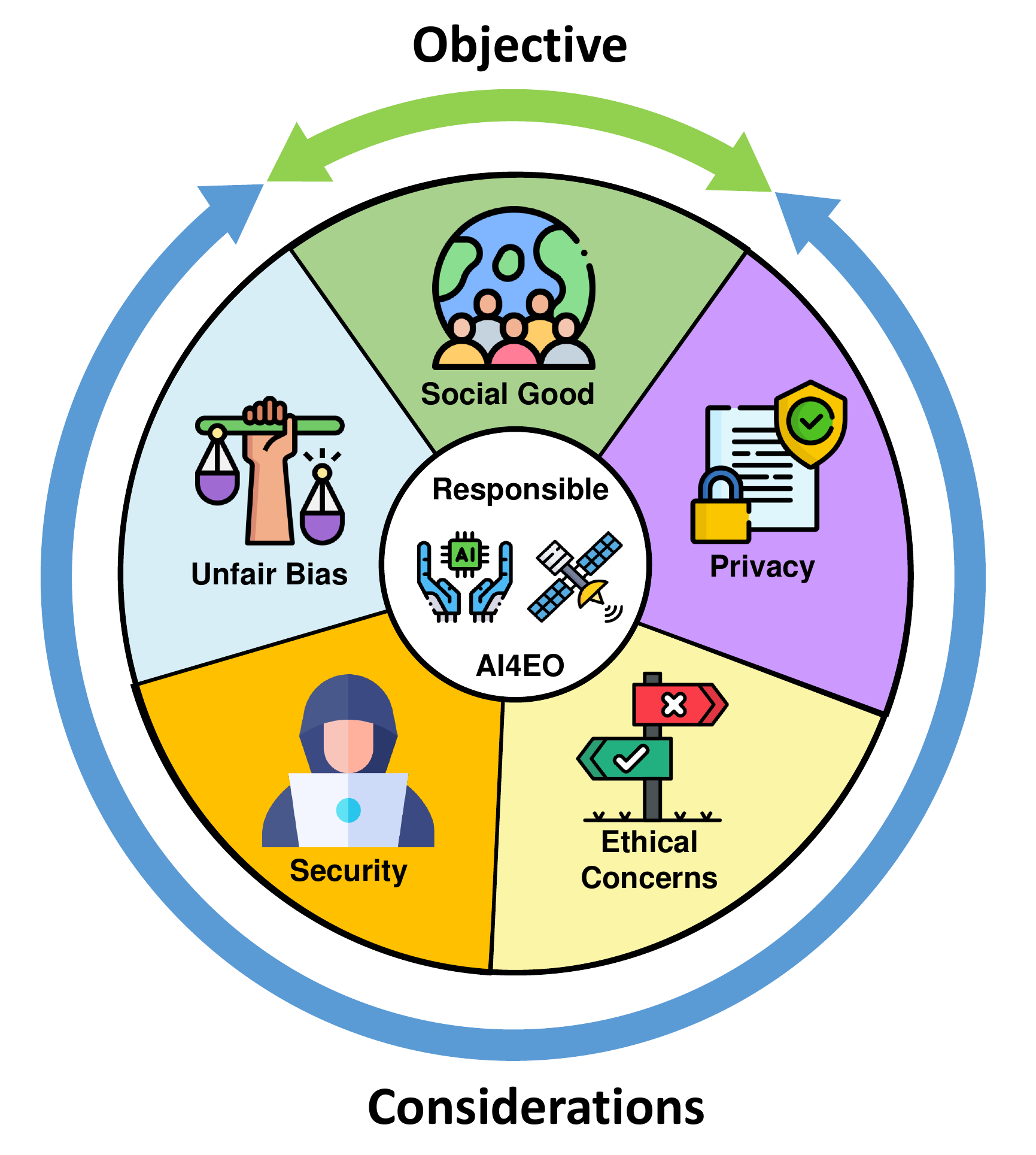}
    \caption{Overview of the main building blocks of Responsible AI in EO: mitigating (unfair) biases, securing AI (defenses, uncertainty modeling, and explainability), preserving (geo)privacy, and addressing ethical concerns outline the considerations necessary for implementing responsible AI methodologies within the fields of EO. Social good presents the opportunities and goals related to how a responsible AI system can effectively be utilized to make a positive difference in people's lives.}
\label{fig:intro_overview}
\end{figure}
Fig. \ref{fig:intro_overview} shows the main considerations of responsible AI in EO, comprising the following five important aspects: 
\begin{itemize}
    \item \textit{Mitigating Unfair Bias (Section \ref{sec:unfair})} involves addressing and minimizing biases present in the algorithms and models used in AI4EO applications. Unfair bias can arise during many stages of the machine learning workflow, such as when the AI systems are trained on data that are not representative or when the data used for training contain inherent biases.
    \item \textit{AI Security in EO (Section \ref{sec:security})} focuses on implementing measures to increase the robustness, control of uncertainty and explainability of the AI systems and the data they process in the context of EO missions.
    \item \textit{Geo-Privacy and Privacy-Reserving Issues (Section \ref{sec:privacy})} aims to find a balance between leveraging the benefits of geospatial data for various EO applications while respecting and safeguarding individuals' privacy rights.
    \item \textit{Ethical Principles in AI4EO (Section \ref{sec:ethical})} introduces the guidelines and standards that govern the maintenance of scientific excellence, data availability and the guidance of AI usage in the context of EO research and applications.
    \item \textit{AI4EO for Social Good (Section \ref{sec:social})} refers to applying cutting-edge AI techniques to address and contribute positively to societal challenges and global issues. It aims to leverage advanced AI4EO technologies to find innovative solutions to pressing problems, promote sustainability, and enhance the overall wellbeing of communities.
\end{itemize}

The main contributions of this paper are as follows: Firstly, we present a pioneering and thorough review of the responsibility of AI research integrated with EO missions for the GRS society. Sections I-IV outline the considerations necessary for implementing responsible AI methodologies within the fields of EO, and Section V presents the opportunities and goals relating to how a responsible AI system can be effectively utilized for social good. This paper emphasizes that responsible AI is not merely a theoretical construct, but an imperative for advancing Earth observation and geosciences. By aligning technological advances with ethical considerations, this discourse aims to contribute to the ongoing dialogue surrounding the responsible and sustainable deployment of AI in the fields of EO and geosciences. Additionally, we dedicate a section (Section \ref{sec: business}) to showcasing the endeavors of private sectors in integrating responsible AI into their business models. This effort aims to foster innovation and sustainability within these sectors. Lastly, Section \ref{sec: conclusion} offers key remarks and actionable guidelines for researchers and practitioners to advance responsible AI practices in EO missions. This review is conducted from the perspectives of both academia and industry, as well as from both theoretical and practical points of view, to ensure a holistic view of the subject. 
	\section{Mitigating (Unfair) Bias}
\label{sec:unfair}


Biases are inherent in all machine learning algorithms. Indeed, courses on machine learning often start with the bias-variance tradeoff, where biases occur when a model is under-fitting and are reduced as model complexity increases. This reduction of model bias is paired with an increase in variance errors as the complexity of the model overfits on variances in training data which are not representative of the target class in general. "Bias" is also a term linked to the systematic errors of an algorithm or dataset. Despite the concept of bias being inherent to machine learning, nowadays there is much concern for biases present in algorithms. Indeed, a review of Responsible AI guidelines and recommendations concluded that the principles of fairness and non-discrimination should be included in all AI guidance documents \cite{fjeld2020principled}. The difference here is that society is particularly concerned about "unfair" biases captured in machine learning algorithms. Generally speaking, this arises where the model systematically produces differing outcomes for specific subgroups of a class in a way that causes societal concern that is "unfair". The unfairness of the model is, therefore, contextual and closely linked to the values of a society. Friedman and Nissenbaum \cite{friedman1996bias}, thus, refer to biases that cause unfair and systematic discrimination of individuals or sub-groups of the population. Preventing discriminatory behavior is also considered to be the responsibility of the developer and user of AI algorithms. The UNESCO Recommendations on Ethical AI adopted by all member states in 2021 states: "AI actors should make all reasonable efforts to minimize and avoid reinforcing or perpetuating discriminatory or biased applications and outcomes throughout the life cycle of the AI system to ensure fairness of such systems" \cite{unesco2021recommendation}. Similarly, the European Commission is drafting an AI Act that incorporates a requirement to consider and report potential biases \cite{euaiact2021}. Understanding how to audit AI workflows for biases and how to mitigate them are expected to be two key fields of research in the coming years.

\begin{figure*}
    \centering
    \includegraphics[width=\linewidth]{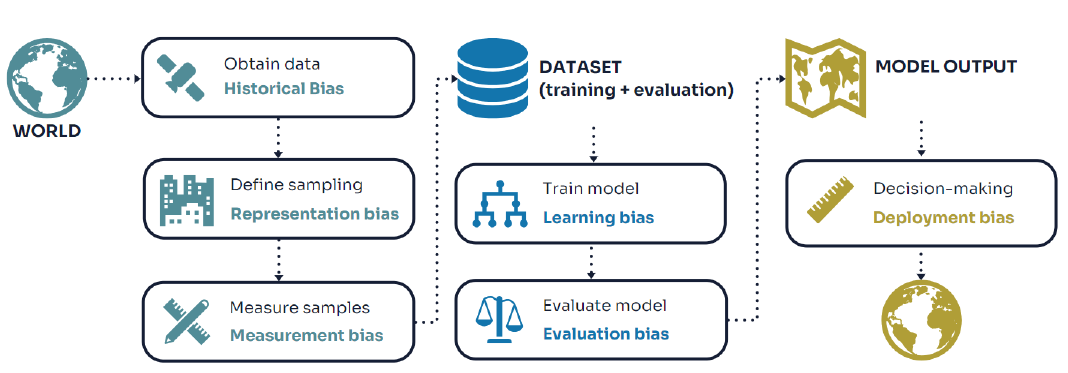}
    \caption{A schematic overview of the different biases that play a role during various stages of a machine learning workflow. Inspired by \cite{suresh2021framework}.}
    \label{fig:biasworkflow}
\end{figure*}

\subsection{Biases in machine learning workflows}
To understand how to identify and mitigate biases, it is important to consider how they emerge at various steps in the machine learning cycle \cite{suresh2021framework,hellstrom2020bias,friedler2019comparative}. Following the framework by Suresh and Guttag \cite{suresh2021framework}, six types of biases that emerge in a machine learning workflow are discussed.
\begin{enumerate}
\item \textbf{Historical bias}, which arises because field data captures phenomena that have already occurred in reality.  If the state of the world changes in the future, the model may not correctly capture those changes. For example, meteorological models trained on historical data may not perform well in the future if extreme events become more prominent \cite{lavell2012climate}. 
\item \textbf{Representation bias}, perhaps the most well-known type of bias when considering fairness. It concerns the way that the real world is sampled to generate the dataset used to train and evaluate the machine learning model. Representation bias occurs when certain parts of the population are under-represented in the sampling.
\item \textbf{Measurement bias}, induced as complex phenomena are defined in simpler concepts during dataset creation. For example, when predicting poverty through EO, poverty is often defined narrowly (e.g., percentage of the population below a defined poverty line) rather than considering broader aspects such as access to housing and sanitation \cite{lucci2018we}.
\item \textbf{Aggregation bias}, where subgroups of a population are aggregated into a single model instead of generating different models for distinct sub-groups.
\item \textbf{Evaluation bias}, stemming from the evaluation set used to assess model performance. Unequal representation of classes in the evaluation set, as well as the use of unsuitable performance metrics, can induce evaluation bias. For example, overall accuracy as an evaluation metric may be unsuitable for classification problems with unbalanced classes in remote sensing.
\item \textbf{Deployment bias}, arising from inappropriate interpretation and usage of the model results for decision-making. This can occur when model results are misinterpreted or applied in contexts for which they were not intended.
\end{enumerate}

\subsection{Auditing and mitigation strategies}
Two main streams of research can be considered in addressing biases in AI workflows: auditing for biases and implementing mitigation measures. 

Auditing a workflow for biases involves identifying causality, such as determining whether a sensitive attribute causes changes in the predicted variable, or examining statistical disparities between accuracy metrics of predictions for sub-groups defined by sensitive attributes \cite{castelnovo2022clarification}. For example, one might investigate whether building detection algorithms perform equally well in high-income and low-income neighborhoods of a city \cite{gevaert2023}. Calculating statistical disparities involves defining a sensitive attribute and determining which fairness metric is appropriate for the application. However, selecting a suitable fairness metric can be challenging as mathematical definitions of fairness may conflict \cite{barocas2019fairness,castelnovo2022clarification}. Interested readers are encouraged to explore resources on similarity-based metrics to understand the differences between metrics and how to select an appropriate one based on the application.

Bias mitigation can be applied at each step of the workflow individually. A recent review on data-centric learning in the field of GRS \cite{roscher2023data} highlights the importance of assessing dataset suitability throughout various stages of machine learning workflows. The review discusses how model bias can be addressed at each of these stages.

\subsection{Challenges and future directions}
There are several challenges specific to detecting biases in GRS applications. 

\subsubsection{Inference of Hidden Sensitive Attributes in ML Models} One challenge lies in defining the sensitive attributes. Societal concerns about which sub-groups should be treated fairly are often connected to demographic and socio-economic characteristics of individuals or groups. The UNESCO Recommendations \cite{unesco2021recommendation} describe fairness according to various categories such as age groups, cultural systems, language groups, persons with disabilities, girls and women, disadvantaged, marginalized and vulnerable populations, rural versus urban areas, and more. Similarly, the EU AI Act \cite{euaiact2021} is founded on Article 21 of the EU Charter of Fundamental Rights, which outlines non-discrimination related to sex, race, color, ethnic or social origin, genetic features, language, religion or belief, minority status, age and nationality, among others. Most of these characteristics are not included explicitly in the data utilized for GRS applications, with the exception perhaps of the urban-rural attribute mentioned by the UNESCO Recommendations. However, indirectly, these characteristics are often inferred. For example, there is extensive research on the identification of slums in satellite sensor imagery \cite{kuffer2016slums, mahabir2018critical}. Although the image does not contain information on poverty directly, the characteristics of the built-up environment have an indirect relation. Similarly, mobility data obtained through mobile phones may overrepresent higher income groups \cite{schlosser2021biases}. Therefore, despite not explicitly listing "sensitive" attributes in the data, geospatial and EO data are implicitly linked to these attributes. A significant challenge is to identify which sensitive attributes could be inferred from the data used to train and evaluate a machine learning model.

\subsubsection{Spatial Correlation Biases in GRS} A second challenge is that of spatial autocorrelation, which is specific to the manipulation of spatial data and is often overlooked when applying methods developed in computer vision directly to geospatial data. Karasiak et al. (2022) demonstrated how the spatial dependence between training and testing samples causes an overestimation of model accuracy \cite{karasiak2022spatial}. Other studies indicated how spatial autocorrelation plays a role in estimating the relations between air pollution and asthma \cite{park2022spatial} and in the estimation of ecosystem services \cite{shaikh2021accounting}. However, auditing for spatial correlation biases is underrepresented in the broader literature on Responsible AI in machine learning. It is expected that identifying and mitigating biases in models will be a particularly active direction of research in the fields of GRS.  
	\section{Secure AI in EO: Focusing on Defense Mechanisms, Uncertainty Modeling and Explainability}
\label{sec:security}

\begin{table*}[h]
\caption{Summary of the 3 AI security concerns including adversarial examples, uncertainty, and explainability.}
\label{table:summary_security}
\centering
\resizebox{\linewidth}{!}{
\begin{tabular}{p{23mm}p{50mm}p{40mm}p{50mm}}
\specialrule{1pt}{0pt}{0pt} 
Security Concerns & Adversarial Examples                                                                 & Uncertainty                                                                 & Explainability                                                                                                   \\  \hline 
Sources           & Model and data-specific adversarial attacks                                          & Error and randomness in the training and testing data, and model parameters & Complexity arising from intricate structures of the \textit{black-box} models.                                              \\
Consequences      & Significantly reduces the performance of trained AI models                           & Inability to control the reliability of model results                       & Untrustworthiness and non-assessability of AI models due to the invisible working process of the \textit{black-box} models. \\
Defense methods   & Adversarial training, randomization, detection, and adversarial purification methods & Uncertainty quantification methods                                          & Explainable AI (XAI) approaches   \\                                              \specialrule{1pt}{0pt}{0pt} 
\end{tabular}}
\end{table*}

In the context of EO, ensuring AI security is vital for upholding responsible AI principles and safeguarding against potential risks and vulnerabilities. EO systems rely increasingly on AI technologies for tasks ranging from image analysis and classification to predictive modeling. However, the integration of AI introduces security concerns such as data integrity threats, adversarial attacks on models, and privacy breaches. Addressing these challenges requires a comprehensive approach that considers both the technical aspects of AI security, such as explainable architectures and uncertainty modeling techniques, and ethical dimensions, including transparency, accountability and the protection of sensitive information. By incorporating AI security measures into the development and deployment of AI4EO systems, we can foster trust, reliability, and societal benefit while mitigating potential harms and ensuring that AI technologies contribute positively to environmental monitoring, disaster response, and sustainable development efforts. While aspects such as ethical dimensions, mitigating unfair biases, and preserving (geo)privacy are largely explained in Sections \ref{sec:unfair}, \ref{sec:privacy}, and \ref{sec:ethical}, we focus on uncertainty modeling, adversarial defenses, and explainability in this section.

\subsection{Security Hazards in AI Algorithms}\label{security-hazards}
Table \ref{table:summary_security} illustrates the primary concerns regarding AI security in GRS, encompassing adversarial attacks, the uncertainty of AI predictions, and the explainability of black-box models. These significant concerns will be addressed in this section. While this paper aims to emphasize concepts, we employ basic equations in this section to define AI security concerns. 

\begin{figure}[ht!]
    \centering
    \includegraphics[scale=0.30]{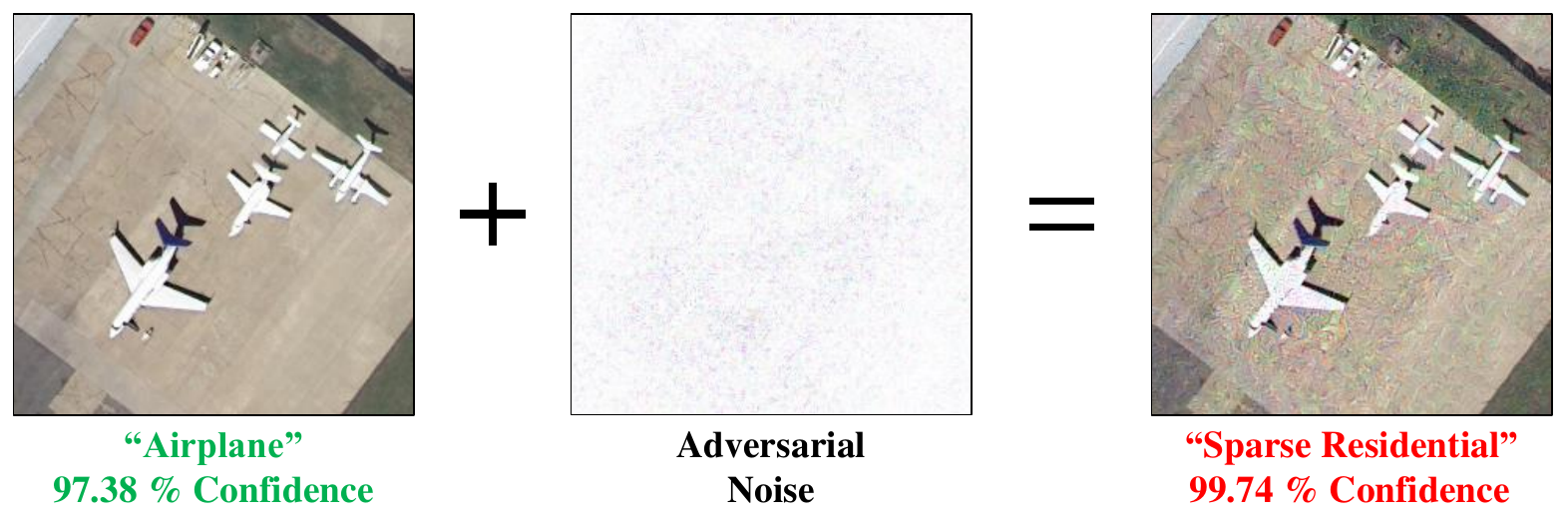}
    \caption{Illustration of adversarial attack: a minor perturbation can fool the classifier into wrong predictions. \cite{yu2023universal}}
    \label{fig:papers}
\end{figure}
\subsubsection{Adversarial Examples}
The identification of adversarial examples in AI4EO tasks \cite{xu2020assessing,chen2021empirical} has forced research into adversarial attacks on deep neural networks (DNNs). By introducing imperceptible perturbations to the data, these adversarial examples can deceive well-trained DNNs, leading to substantial performance degradation, even in state-of-the-art AI approaches like image classification and semantic segmentation \cite{xu2022universal}. Since the deep learning-based AI4EO methods minimize loss functions as an object to train the models, adversarial attacks employ the idea of maximizing the loss functions to find a perturbation to contaminate the clean samples, which can consequently fool well-trained AI4EO algorithms. As a result, this behavior severely threatens the responsibility of the current AI4EO approaches by making them predict incorrect results with high confidence. 

Adversarial attacks can appear virtually and physically when applying deep learning models.
On the one hand, the attackers can directly modify the applied remote sensing data before they are forwarded to the application models. For example, \citet{wu2023adversarial} demonstrated that the current end-to-end autonomous driving system could be easily broken by tempering the data with mild perturbations generated by white-box attacks. On the other hand, physical attack methods are more imperceptible since the adversarial perturbations are introduced during the remote sensing imaging process \cite{czaja2018adversarial}. \citet{zhang2022adversarial} explored the physical attack in real-world AI4EO applications by applying an adversarial patch attack on the multi-scale object detection for UAV remote sensing images.
As a corollary, responsible AI is required to mitigate the potential threats of these adversarial perturbations. Adversarial defense methods are desired that improve the reliability and response of the model, which are introduced in a later paragraph in this section.

\subsubsection{Uncertainty of the Model Predictions}\label{uncertain}
GRS data, often sourced from varied platforms, exhibit diverse domain distributions, introducing uncertainty in AI model predictions, as described in Section I. The stochastic nature of machine learning models further amplifies this uncertainty, posing a significant challenge to prediction reliability \cite{foody2002current}. Recognizing and managing uncertainty is vital for the development of responsible AI4EO models. The application of AI in GRS data involves distinct training and testing phases to optimize a mapping function $f_{\theta}: \mathcal{X}\to \mathcal{Y}$, where $\theta$ represents the initialized parameters. To this end, uncertainties may arise in the samples of training and testing datasets, as well as in the model parameters, categorized into data and model uncertainty \cite{dungan2002toward,persson2020characterizing}.

Lack of domain knowledge is a prominent factor contributing to data uncertainty, arising from the disparate domain distribution observed between the training dataset $\mathcal{X}$ and the testing dataset $\mathcal{X}^{*}$. In remote sensing imaging, spectral features of observed data are closely tied to spatial and temporal conditions, such as solar illumination, season, and weather. Consequently, variations in these conditions can markedly influence imaging outcomes, resulting in heterogeneous data with diverse domain distributions, a phenomenon known as domain invariance \cite{tuia2016domain}. Regrettably, existing AI approaches struggle to make accurate predictions with a decision boundary that accounts for data exhibiting domain invariance, often referred to as domain shift \cite{tuia2021recent}. This poses a significant threat to model performance, particularly when faced with data uncertainty in testing samples during the model deployment stage. In attempts to mitigate domain shift issues, numerous studies in geoscience and remote sensing have proposed methods to adapt and extend the decision boundary of classification models to encompass unknown data in inference samples \cite{zhang2023cross,xu2022eyes}. However, these methods can only fine-tune models to a limited extent, and the data uncertainty associated with domain invariance cannot be eradicated entirely.
In addition to data uncertainty, model uncertainty encompasses the errors and randomness inherent in the model parameters $\theta$, initialized and optimized during the model design and training processes. As researchers strive to develop advanced AI algorithms with cutting-edge performance and efficient training, diverse model architectures linked with optimization strategies have been devised for various GRS applications \cite{lin2021comparative}. However, determining the optimal model architecture and training hyperparameters remains challenging, introducing uncertainty into the trained model that can subsequently impact predictions, commonly referred to as model uncertainty \cite{doicu2021overview}.


\subsubsection{Opacity of Black Box Models}\label{blackbox}

As high-performance computing technology advances, contemporary AI applications in geoscience and remote sensing studies explore models with deeper architectures and more complex parameters to enhance classification performance. However, the increased complexity raises challenges in understanding and interpreting the internal operational steps of these architectures, rendering the models akin to black boxes \cite{black2018model}. Compared with traditional machine learning approaches, the opacity of these black box models hinders the identification of hidden security hazards that can significantly impact classification, as well as the reasons for "wrong" predictions in real-world applications \cite{brendel2017decision}.
Moreover, the growing emphasis on data privacy and high confidentiality in many GRS tasks underscores the need for trustworthy AI models aligned with ethical and judicial requirements for both developers and users.


\subsection{Robust Modelling for AI Security}

In response to the identified security hazards outlined in the preceding section, researchers are working towards the development of more robust AI models, emphasizing enhanced responsibility for AI security. Several strategies are being employed, including adversarial defense, uncertainty quantification and explainable AI, to address challenges related to adversarial examples, data and model uncertainty, and black-box models, respectively.

\subsubsection{Adversarial Defense}

Common adversarial defense approaches include adversarial training, randomization, detection, and adversarial purification methods. Adversarial training methods aim to bolster the model's robustness against adversarial examples by incorporating them directly into the model training stage alongside normal training samples. For example, Goodfellow \cite{goodfellow2014explaining} proposed an empirical adversarial training scheme involving the inclusion of extra-generated Fast Gradient Sign Method (FGSM) attack adversarial samples. However, it is important to note that adversarial training methods may enhance the model's robustness against specific attack methods, yet vulnerabilities against adversarial examples of other types may persist.

Given that adversarial perturbations can be perceived as additive noise on adversarial examples, randomness methods introduce random components to enhance the model's robustness. For example, Cohen developed a randomized smoothing technique for adversarially robust classification involving the randomization of the input to the DNN to eliminate potential adversarial perturbations in the input samples \cite{sitawarin2022demystifying}. However, it is important to note that the performance of randomness methods depends greatly on the quality of the randomization process and can be affected significantly by both the lack of prior knowledge of attack types and theoretical explanations.


\subsubsection{Uncertainty Quantification}

To better comprehend and manage potential uncertainty, uncertainty quantification techniques have been proposed to assess the credibility of predictions.

In many GRS applications, classification results manifest as distributions of possibilities associated with object classes. These distributions are typically transformed by a softmax function at the final stage of the network. Grounded in the principles of Bayesian inference, the data uncertainty of an input sample $x^{*}$ can be described as a posterior distribution over the class label $y^{*}$ given a set of model parameters $\theta$. Similarly, the model uncertainty can be formalized as a posterior distribution given the training dataset $\mathcal{D}=(\mathcal{X},\mathcal{Y})$, as expressed by the equation:
\begin{equation}
	\mathcal{P}(y|x^{*},\mathcal{D})=\int  \underset{\mathrm{data}}{\underbrace{\mathcal{P}(y|x^{*},\theta)}}\ 
	\underset{\mathrm{model}}{\underbrace{\mathcal{P}(\theta|\mathcal{D})d\theta}}.
\end{equation}
Building upon this equation, various uncertainty quantification approaches aim to obtain an uncertainty distribution $\sigma^{*}$ of model predictions $y^{*}$ by marginalizing $\theta$. These methods fall into two broad categories: deterministic methods and Bayesian inference methods, differing in subcomponent model structures and error characteristics. In deterministic methods, the parameters of AI models are deterministic and remain fixed during the inference step, aligning with the common behavior of most developed AI models. In contrast, Bayesian inference methods employ the Bayesian learning strategy, leveraging the ability to seamlessly integrate scalability, expressiveness, and predictive performance of neural networks for uncertainty quantification \cite{ma2021corn,de2014uncertainty}.

\subsubsection{Explainable AI}
\begin{figure}[ht!]
    \centering
    \includegraphics[scale=0.20]{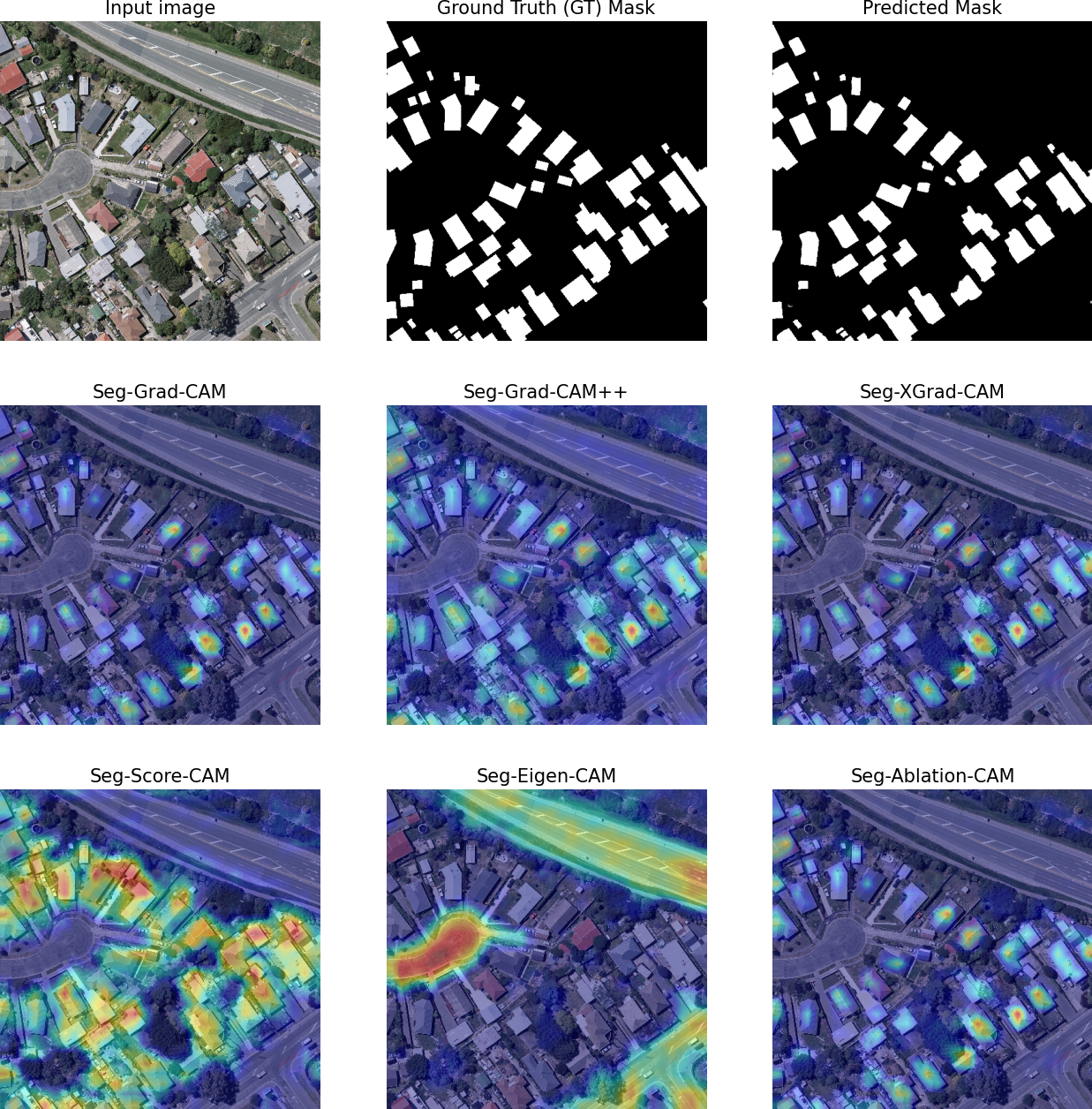}
    \label{fig:XAI}
    \caption{Heatmaps of UNet decoder block with different CAM-based XAI algorithms for the class of "buildings" in the semantic segmentation of an RS image. (source from \cite{gizzini2023extending}). Pixels that have a brighter color indicate a higher probability of being classified into the target class.}
\end{figure}

In general, explainable AI (XAI) aims to provide an understanding of the pathways through which output decisions are made based on the parameters and activations of the trained models. To this end, many XAI approaches have been proposed for a comprehensive understanding of the model behavior to justify the individual feature attributions of a test sample $x \in \mathcal{X}$, which provides a powerful tool to examine incorrect predictions and inspect the potential threats of black-box models, especially in high-risk GRS applications \cite{hasanpour2022unboxing}. For example, \citet{ribeiro2016should} developed a Local Interpretable Model-agnostic Explanation (LIME) method to provide understandable representations for the predictions of black-box classifiers by highlighting attentive contiguous superpixels of the source image with positive weight towards a class, which benefits users in learning "how the model thinks" when classifying a specific image. Furthermore, the deep features of DNNs can also be processed into visualizations to observe how the training procedure works on specific tasks, as shown in Fig. \ref{fig:XAI}.

\subsection{Challenges and Future Perspectives}
In this section, we discuss the future perspectives of AI security for EO tasks from three perspectives:

\subsubsection{The Game between Adversarial Attack and Defense towards EO Applications}
While adversarial examples have been identified widely in several EO applications, research on adversarial attacks and defense algorithms has been developed and raised the debate in game theory. On the one hand, the attackers have proposed more powerful attack algorithms to affect a wider range of DNNs applied in various EO missions, including scene classification and semantic segmentation. On the other hand, the defense group has also identified the emerging adversarial threats and tended to secure AI algorithms from more complex adversarial examples with frontier techniques, such as diffusion models \cite{yu2023universal}. In conclusion, the game between adversarial attack and defense algorithms has been started and extended to more subtopics in the AI4EO research field \cite{pal2020game}.

\subsubsection{Practical Uncertainty Analysis for AI4EO}
The uncertainty in both model and data towards AI4EO missions has been a critical obstacle for developing AI models with more reliability and responsibility. Although some uncertainty quantification methods have been seen in geoscience and RS studies, most of them were designed for dedicated models and are not adaptable to most of the existing AI4EO approaches. To this end, universal uncertainty analysis with greater capability to be applied in more typical DNNs should be developed for the AI4EO missions.

\subsubsection{XAI Schemes with more Understandable Interpretations}
Explainability is considered to be a significant attribute of modern AI4EO missions, which provides the opportunity for humans to learn the intrinsic process of the AI algorithms, security risks can be largely alleviated. However, current XAI methods designed for EO missions can only generate visualization interpretations, which are usually semantically misaligned and still too abstract to understand directly. Therefore, future XAI studies should focus on increasing the depth and accuracy of XAI results by introducing more constraints and optimization problems to explanations.

	%
	%
	\section{Geo-Privacy and Privacy-preserving Measures}
\label{sec:privacy}
  

Remotely sensed images and geospatial data offer unprecedented opportunities to monitor our planet and support spatial planning, management and decision-making. Nevertheless, as remote sensing technologies evolve, the need to balance the advantages of data-driven decision-making with the protection of individual privacy becomes paramount. The increasing resolution and quality of remotely sensed data, with their ability to capture detailed information about the Earth's surface and (directly or indirectly) the people living in those areas, raises concerns about the potential misuse of data and the accidental revelation of sensitive information \cite{HARRIS2021101422}. AI technologies allow the scale-up of geospatial applications, reducing analysis costs and subjectivity and, at the same time, magnifying the risk of exposing sensitive personal information.

Effectively addressing the above challenges requires thoughtful consideration of ethical principles, legal frameworks, and technological safeguards to ensure that the benefits of geospatial advances are realized without compromising individual privacy rights. Earlier research endeavors aimed at constructing a conceptual privacy framework; for example, \cite{finn2013seven} identified seven different concepts of privacy by examining distinct emerging technologies. 

\subsection{Privacy in UAV images}
In remote sensing, privacy concerns often arise when the resolution of an image is sufficiently fine to discern individuals' identities and expose personal details. This scenario commonly occurs during the analysis of images captured by Unmanned Aerial Vehicles (UAVs), potentially encroaching upon: (1) Location privacy, as individuals can be identified and tracked within UAV images. (2) Behavior privacy within private settings, free from external surveillance. (3) Space privacy, revealing details about secluded areas like backyards. (4) Association privacy, disclosing group memberships and affiliations. (5) Data and image privacy, involving individuals' control over images featuring their presence. 

Gevaert et al. \cite{gevaert2018evaluating} evaluated the societal impacts of using UAVs for informal settlement mapping through two case studies in Eastern Africa. The study stresses the importance of notifying citizens of the purpose of data collection, the rights to access, data processing, and distribution, especially when flying over marginalized communities. The absence of a cohesive policy framework governing such activities shifts a significant portion of responsibility onto industry self-regulation, potentially falling short of adequately safeguarding marginalized communities. Nevertheless, the existence of a ‘unified’ framework is challenged by the specific application of UAV image acquisition and the fact that the concept of sensitive information or privacy may vary among people, groups, and cultures. The study illustrates the importance of considering the local context regarding privacy issues.
Another example application with potentially privacy-related concerns is the use of UAV images in support of first responders in emergency situations (e.g., after an earthquake or other types of disasters). The authors in \citep{zhang2022training} developed a deep-learning network to detect victims in UAV images, which is trained to detect humans even in complex situations where body parts of the victims are partially covered by dust or debris. However, collecting a dataset for training such a network is challenging because of the sensitive nature of the data. Therefore, the authors proposed a framework to generate a harmonious composite of images for training. The framework first pastes images of body parts onto a debris background to generate composite victim images, then uses a deep harmonization network to make the composite images more harmonious.

\subsection{Mapping Sensitive Information}
Besides decimeter/centimetre-resolution UAV images, privacy concerns may also arise with satellite sensor images of various resolutions, especially when applied to map societal and socioeconomic features. Examples of sensitive applications include the mapping of informal settlements, slums,  deprived urban areas \cite{kuffer2018scope, kuffer2021spatial, persello2017deep} and refugee camps \cite{HASSAN2022103120}. The peril associated with disclosing sensitive information is evident in the "Mass Atrocity Remote Sensing" initiative, which sought to furnish forensic evidence of crimes against humanity through the analysis of satellite sensor imagery and other datasets. Nevertheless, researchers soon recognized that these data could also be exploited to identify and harm individuals and groups.
Publishing sensitive maps should always be done with care to prevent unintended consequences for local communities. In the case of informal settlements, being classified as living in such a neighborhood may result in social exclusion or even exposure to eviction (e.g., see the reports of forced evictions of 30,000 people from Nairobi’s largest slum, Kibera, in 2018 for building a new road \cite{bbc2018kenya}). Measures for privacy preservation typically include downgrading map spatial resolution (e.g., to cells of 100×100 m), aggregating data at the administrative unit level, and removing geo-location from datasets.

Another strategy to preserve household privacy is to apply jitter to geospatial point data (adding random noise to geocoordinates). The authors of \cite{chi2022microestimates} developed a comprehensive and openly accessible collection of microestimates detailing the distribution of relative poverty and wealth across 135 low- and middle-income countries (LMICs). They gathered data from Demographic and Health Surveys (DHSs) through traditional face-to-face surveys conducted with 1,457,315 distinct households residing in 66,819 villages spread across 56 different LMICs globally. The DHS data contain approximate GPS coordinates indicating the centroid of each surveyed village among the 66,819. However, to safeguard privacy, the precise geocoordinates are obfuscated, with a margin of error of up to 2 km in urban areas and up to 5 km in rural areas.

\subsection{Challenges and Future Perspectives}

A recent challenge is the increasing availability of satellite sensor images with sub-meter resolution up to 30 cm (e.g., WorldView-3, Pleiades NEO, Skysat) able to capture potentially sensitive information. The potential of such data for mapping and object detection should, therefore, be counterbalanced with privacy-preserving measures. Moreover, it is worth noting that privacy requirements may clash with open science principles, restricting the possibility of openly publishing and benchmarking datasets. 

Future investigations should develop strategies to openly share data for scientific advancements while preserving sensitive personal or group information. Such strategies are much needed in applications such as mapping deprived urban areas or victim detection, where further advances in the development of AI-based methods critically depend on the availability of large benchmark datasets, which are currently missing.






 
	%
	\section{Maintaining Scientific Excellence, Open Data, and Guiding AI Usage Based on Ethical Principles in EO}
\label{sec:ethical}
  

The real-world character of AI4EO applications with downstream political, societal, and environmental impacts offers tremendous opportunities for positive change while simultaneously bearing the risk of negative impacts when not carefully designed and implemented \cite{hanson_garbage_2023, rolf_mission_2024, wagstaff_machine_2012}. Maintaining the highest scientific standards and making data and code openly accessible while complying with ethical principles are essential prerequisites to enhance the transparency of knowledge production, ensuring high reproducibility of scientific work and maximizing the societal, economic or environmental benefits of AI4EO research, development and applications \cite{hanson_garbage_2023}. Compliance with fundamental ethical principles is, thus, a major duty of the scientific community.  

A series of review papers laid out the relevant definitions and concepts of ethics in AI. For example, Jobin et al. (2019) \cite{jobin_global_2019} identified five broad categories from ethical guidelines in AI research: transparency, justice and fairness, non-maleficence, responsibility and privacy. With a focus on AI4EO, Kochupillai et al. (2022) \cite{kochupillai_earth_2022}, layed out six ethical dimensions: privacy, honesty, integrity, fairness, responsibility, and sustainability (for an overview on recent AI ethics categorizations, see Table 2 in \cite{kochupillai_earth_2022}). We here revisit the six categories proposed by \cite{kochupillai_earth_2022} focusing on aspects that transcend the themes of maintaining scientific excellence, open data, and guiding AI usage in AI4EO, that is, directly or indirectly relate to workflows, from defining objectives and research questions, data inputs and analysis outputs and their downstream impacts.  

\subsection{Privacy}\label{eth_privacy}

In the context of responsible AI4EO, we situate privacy concerns arising under the open data paradigm and with regard to (territorial) stigmatization. Open access to datasets and code is a core element of scientific reproducibility. Scientific excellence in AI4EO can, thus, only be achieved when the public dissemination of essential research data and code are made available to enable full reproducibility and robustness checks \cite{kedron_how_2022}. The scientific community has pushed for improving the Findability, Accessibility, Interoperability and Reuse of digital assets, which materialized in the FAIR-principles \cite{wilkinson_fair_2016}. Increasingly, however, the publication of geospatial data, including UAV or fine to very-fine-resolution satellite sensor imagery, labeled datasets, or the outputs of AI4EO workflows, may raise privacy concerns \cite{richardson_replication_2015}, see \ref{sec:privacy}). The FAIR-principles demand data to be free from use restrictions, fees or embargos, but legal, moral, or ethical reasons, such as safeguarding individual privacy, or endangered species, may justify deviation from these best practices on data accessibility \cite{kinkade_geoscience_2022}. However, detailed metadata should be made available enabling the discovery of such data and describe explicitly the conditions under which data access can be granted. 

Trade-offs between the benefits of open data and privacy concerns therefore need to be evaluated case-by-case to avoid privacy-violating or other ethically concerning impacts of public data dissemination. Besides the techniques to obscur geolocation information described above, a selective disclosure of data upon requests may be justified to avoid risks of data misuse. Additionally, purpose-specific data licensing may provide opportunities to avoid unintended use of datasets. One example for purpose-oriented data releases is Norway's International Climate and Forests Initiative (NICFI) Satellite Data Program, providing commercial satellite imagery covering the world´s tropics for non-commercial sustainability-centered applications (https://www.planet.com/nicfi/).

Stigmatization may occur in cases where urban neighborhoods are categorized as slums \cite{kochupillai_conducting_2023, oluoch_crossing_2024}, or categorizations of crime rates or poverty levels adversely affect public perceptions of a community \cite{tang_predicting_2022}. The resulting territorial stigmatization may influence decision-making and negatively affect the local population. Similarly, stigmatization applies to AI4EO applications failing to consider the realities on the ground, for example, designed to expose illegal activities, such as deforestation, fire use, or mining activities, without knowledge about potential concessions to conduct the respective activity. Similarly, stigmatization may occur when global or broad-scale analyses fail to sufficiently consider local realities. For example, small-scale farming has been found to be responsible for 97\% of deforestation on the African continent \cite{branthomme_how_2023}, but regional dynamics such as the rapid emergence of medium-scale commercial farming operations \cite{jayne_changing_2022, wineman_relationship_2022}, the increasing role of export-oriented commodity crops such as cocoa, cashew and oil palm in deforestation \cite{masolele_mapping_2024}, and trade-offs between environmental degradation, food production, livelihoods and labor opportunities in semi-subsistence agriculture \cite{chiarella_balancing_2023,meyfroidt_trade-offs_2018} allow for a more nuanced perspective of such dynamics. As such, avoiding stigmatization of people, communities, or localities requires a high level of awareness of researchers in the design of research questions and label heuristics, as well as embedding scientific findings and their limitations in the realities of the region under investigation \cite{tulbure_regional_2022}. We highlight that cooperation with regionally knowledgeable partners in the early project design phase is an efficient way to avoid pitfalls of stigmatization in responsible AI4EO \cite{haelewaters_ten_2021}.
 
\subsection{Honesty}\label{eth_honesty}

Truthfulness and trustworthiness are inherent to scientific inquiry, but the current replication crisis undermines the credibility of science and, thus, hampers the societal uptake of scientific outputs. As a community producing knowledge with real-world impact, responsible AI4EO requires transparency, explainability, and data veracity \cite{hanson_garbage_2023}. 

In AI4EO, transparency not only requires thorough documentation and referencing of the datasets and workflows used but also that need to understand and clearly communicate ambiguities, potential biases, and errors, or conceptual limitations. In light of the ongoing democratization of EO data and workflows through cloud platforms, and accessible user interfaces, empowerment through access to EO data requires a thorough understanding of the limitations of EO data and workflows. Furthermore, transparency enables critical reflection of scientific findings. One example is a recent study linking the observed emerging abilities of LLMs (sudden, unexpected increases in performance) to study design choices made by researchers (e.g. evaluation metrics) \cite{schaeffer_are_2023}. 

Assuring explainability in AI4EO continues to be challenging, but is a prerequisite for early diagnostics regarding unwanted bias and sources of error. We strongly encourage the community to conduct  research on XAI, and incorporate it, wherever possible. Focusing research on explainable AI4EO will contribute to the recognition of EO data as a distinct modality in the AI community at large, by better understanding how AI models leverage the information contained in EO data \cite{rolf_mission_2024}. Greater explainability may also profit the emerging sub-field of self-supervised learning \cite{wang_self-supervised_2022}, and help to ensure oversight of (geographic) biases in AI4EO foundation models which claim high degrees of geographic generalization, but which are trained in selected world regions \cite{jakubik_foundation_2023}. Such biases are, for example, present in LLMs \cite{manvi_large_2024}, and may lead to error propagation to many downstream applications \cite{bommasani_opportunities_2021}. 

Data veracity in EO concerns both the imagery and the EO data pre-processing workflows used, as well as reference data / labels being in accordance with a (non-stigmatizing) heuristic defining the objects of interest. Using well-established and tested processing pipelines to assure the highest possible quality of EO data, for example, in terms of atmospheric correction \cite{doxani_atmospheric_2023}, but also regarding QA flags, including cloud masks for optical data \cite{skakun_cloud_2022}, is recommended to avoid downstream errors related to poor data quality. Labeling (the process of categorizing real-world phenomena into distinct classes or groups) is highly context-specific. Therefore, it requires regional expertise and knowledge, is ideally coordinated with local experts, and can be considered an iterative process in which exchange of knowledge from various disciplinary lenses is an effective means to reduce the risk of false categorizations or stigmatization \cite{tulbure_regional_2022}.

\subsection{Integrity}\label{eth_integrity}
Technical robustness and uncertainty reporting are key elements of integrity in AI4EO. Detailed documentation and reporting of the employed processing pipelines, including the respective processor version, enhances the reproducibility of AI4EO analyses. Detailed reporting of biases, errors, and uncertainty adds credibility and is detrimental to avoid erroneous use of datasets in downstream applications. Given the increasingly close interactions between the ML and EO communities, the maintenance and use of well-established evaluation standards particular to the EO domain, such as in reporting map accuracies and area estimates in categorical settings \cite{olofsson_good_5}, or for detecting trends in spatiotemporal data \cite{ives_statistical_2021} is key. 

In AI4EO research, a focus on single aggregate evaluation metrics (e.g. overall accuracy), as discussed in Section \ref{sec:unfair}, is insufficient, as often the detection of minority classes is most relevant in many AI4EO applications. Generally, and for imbalanced problems in particular, reporting area-adjusted class-specific user´s and producer´s accuracies from an independent and randomized sample must be preferred over simplistic one-dimensional statistics. Similarly, class area estimates are to be derived from reference samples rather than pixel counting in maps with non-random error distribution. For continuous targets (e.g. reflectance, biomass, crop yields, canopy height), reporting multiple error metrics and assessing them across the entire value range reveals heteroskedasticity and can inform downstream users of potential pitfalls when using a product \cite{frantz_understanding_2023}.

\subsection{Fairness}\label{eth_fairness}

Fairness here refers to avoiding bias, and to respecting all aspects of diversity and sociocultural differences, ultimately creating standards based on principles related to fairness. Bias may relate to all stages of AI4EO analyses, beginning with defining a research question, through the workflow itself, and ending with the manifold downstream impacts related to interpreting the results - as laid out in \ref{sec:unfair}. We will not delve deeper into the broader ethical framework discussed elsewhere \cite{kochupillai_earth_2022, xu_ai_2023}. Instead, we focus on core needs, before, during and after setting up an AI4EO workflow.

Before setting up an AI4EO workflow, we may consider opportunities to invite potential stakeholders of anticipated outcomes to participate in co-creating the research agenda  \cite{brandsen_definitions_2018}. Stakeholders typically include different institutions across all levels, from village heads in an area of interest to the United Nations or the World Bank. Stakeholders may also be citizens who are acquainted with the topic or the area of interest, ideally (but not mandatory) being engaged in related citizen-science opportunities \cite{verburg_land_2015}, such as data sampling or field-based mapping evaluation \cite{fritz_geo-wiki_5, schepaschenko_development_6}. Citizens and institutions acquainted with the research questions will help to mitigate bias in the project design phase.

During workflow creation, interpretable AI, as discussed in Section \ref{sec:security}, should support paving the way towards optimizing workflows and avoiding downstream misinterpretations \cite{gevaert_explainable_2022, rudin_stop_2019}. Interpretable AI also allows tuning algorithms and workflows based on the scientists’ and non-science stakeholders’ expertise \cite{graziani_global_2023}. Non-scientists thereby provide consultancy to EO scientists to overcome the different biases mentioned in Section \ref{sec:unfair} during the workflow implementation and roll-out, including historical, representation, measurement, aggregation and evaluation bias (compare \ref{sec:unfair} A). Next to training data creation, providing sufficient and independent evaluation data for accuracy assessment is both of core importance and often one of the most challenging tasks (see also \ref{eth_integrity}).

Implicitly, avoiding workflow-related bias is also tightly connected to recognizing deployment bias.  Deployment bias relates to the inappropriate interpretation of results after the data analysis, which is often related to downstream decision-makers. This underlines the importance of co-creation and co-production to minimize and mitigate deployment bias early on. In other words, confronting stakeholders with results created in the scientific “ivory tower” without engaging with them in the design and production phase regularly creates unsatisfactory results for those affected or supposed to build their decision-making processes on AI4EO outcomes. 

Deployment bias also points directly to the need to respect diversity and sociocultural differences. For example, we need to recognize different ethnic or religious communities, as well as AI4EO experts compared to non-expert stakeholders to avoid mis-interpretations or disrespectful outcomes of AI4EO workflows. Typical examples are the mapping of slums or of agricultural practices in the Global South \cite{kochupillai_conducting_2023, nakalembe_considerations_2023}. Many studies on related research questions need to integrate stakeholders early on in AI4EO to include the necessary expertise to create meaningful results, specifically across large areas of interest, where diverse and multicultural stakeholders are needed. AI4EO workflows also rarely consider resource restrictions in the Global South - regarding the “data desert”, limited processing resources, or limited financial backup, e.g. for implementing cloud computing or using HPC or supercomputing facilities \cite{stahl_ethics_2023}. In other words, cooperation at “eye level” is needed to embed AI4EO in the proper sociocultural context, specifically when solving SDG-related questions and global change issues in the Global South \cite{gevaert_explainable_2022, haelewaters_ten_2021}. In conclusion, we create win-win situations when joining the forces of stakeholders "on the ground" (deliberately including stakeholders across scales from local dwellers to global institutions, depending on the research questions to be tackled) with AI4EO experts.

\subsection{Responsibility}\label{eth_responsibility}
The need for interdisciplinary and regional collaboration related to domain knowledge across the entire workflow of AI4EO and unwanted outcomes or interpretations were tackled in the section on “Fairness”. We, therefore, here focus on handling the distinctness of EO data well \cite{rolf_mission_2024}. 

It has been argued that the AI/ML community urgently needs to recognize the distinct features of satellite-based EO data and to “move out of the comfort zone” of applying AI methods established in other domains without improved adaptation to EO or satellite data machine learning (SatML, \cite{rolf_mission_2024}). The reasons are manifold as, for example:
\begin{itemize}
    \item Satellite data is rarely “analysis-ready data” (ARD) \cite{frantz_forcelandsat_2019}), requiring state-of-the-art preprocessing to represent the same values for the same features in different images, i.e., to avoid pseudo-variance.
    
    \item Spatial autocorrelation according to Tobler’s first law of geography \cite{miller_toblers_2004} is rarely used to improve AI4EO, even less so newer concepts like Partitioned Autoregressive Time-Series (PARTS) that consider both spatial and temporal autocorrelation in a congruent fashion \cite{ives_statistical_2021}

    \item Satellite data provide a wide range of spatial resolutions (from very-fine-resolution to pixels of several $\mathrm{km}^{2}$) and extend across all scales to be observed (from local, fine-scale objects in urban environments to global phenomena). Both very diverse spatial detail and multi-scale research ask for adapted solutions.
    
    \item Different temporal resolutions, i.e., data densities over time, represent the same land surface differently. In other words, change analyses over years or decades are far from trivial, and need contextualization \cite{frantz_revisiting_2022, frantz_understanding_2023}. This is even more important when phenology is considered and specific temporal windows across the seasons are crucial to identify the relevant phenomena, e.g. different crop types, land use intensities or tree species separable only at certain phenological stages \cite{blickensdorfer_national_2024,lobert_deep_2023,schwieder_mapping_2022}.

    \item Most AI algorithms for image analyses were developed on three-band red-green-blue (RGB) imagery, often based on VHR drone data. Today’s multi- to hyperspectral satellite data ask for more sophisticated approaches to detect not just the obvious, but the relevant. This is aggravated by more spectral detail in radiometric data beyond 8-bit and even more by using virtual constellations \cite{wulder_virtual_12}, ranging from optical over radar to LiDAR data.

    \item Sparse labeling problems occur with increasing areas of interest enabled by AI4EO. More severe problems are inherited by biased samples across larger regions or non-trivial features to be sampled from existing data or the imagery itself \cite{rufin_taking_2023, wang_unlocking_2022}.
\end{itemize}

While all these aspects are relevant for any EO analysis framework, opportunities have increased rapidly with AI4EO (e.g., regarding the size of areas of interest that can be handled). Along with these are the related risks and potential awareness deficits. Many, if not the majority of, AI4EO studies focused on comparably small use cases in the past to develop or apply algorithms. However, the roll-out to real-world problems, for example, at the national, continental or even global scale, requires ubiquitous algorithms that can handle much more diverse feature spaces (over time). There is indeed a lack of upscaling studies demonstrating the suitability of most algorithms for tackling real-world problems \cite{rolf_mission_2024, tuia_toward_2021, tuia_artificial_2023}.

\subsection{Sustainability}\label{eth_sustainability}
Since 2015, the United Nations has set an ambitious agenda related to the Sustainable Development Goals (SDG), aiming to improve social, economic and environmental conditions globally by 2030 \cite{un_transforming_2015, Persello_2022}, with a wealth of implications for AI and also specifically for AI4EO \cite{stahl_ethics_2023}. Here, we take the broadest possible standpoint, including the intrinsic sustainability of AI4EO workflows and the external impacts for all three SDG dimensions. Given that AI at large offers the potential for both positive and negative impacts across all SDGs, with a majority on social sustainability \cite{vinuesa_role_2020}, AI4EO has a similar potential for both positive and negative impacts.

Focusing on environmental sustainability, EO is data- and energy-hungry, thereby creating immense $\mathrm{CO}_{2}$ emissions. This footprint relates to the entire lifecycle of EO engineering, science and outcomes, including but not being limited to building, launching and running satellite missions, data creation and curation, data storage and data analysis workflows - the data-related aspects being core for AI4EO \cite{wilkinson_environmental_2024}.

Zooming in on economic and social SDG, the FAIR principles are highly relevant: We thematized avoiding “helicopter research” already in \ref{eth_fairness} \cite{haelewaters_ten_2021}. Offering inclusiveness is non-trivial to deliver, as many EO scientists and companies, specifically in the Global South, do not have access to powerful cloud computing or HPC services or to GPU-equipped high-performance workstations. Even if access to online services may become available through dedicated programs or international cooperations, proper education and training for the next generation of AI4EO scientists is of utmost importance \cite{stahl_ethics_2023}.

The outcomes from AI4EO research may greatly support improvements along the lines of the UN SDGs \cite{Persello_2022, vinuesa_role_2020} - but may also lead to ill-posed conclusions. Such conclusions can, in the worst case, create an existential economic threat (e.g., if loans or insurance are not provided based on AI4EO-derived indicators \cite{benami_uniting_2021}). Also, local stakeholders including indigenous people, who may not even be aware of governments’ decision making processes, are regularly impacted, though. Last but not least, AI4EO may be in the “wrong hands”; for example, used by autocrats, dictators, or criminals to suppress or threaten specifically environmental and social SDGs. The most disturbing examples of this are ubiquitous AI4EO-supported surveillance and warfare.

	\section{AI\&EO for Social Good}
\label{sec:social}

In this section, we build upon the discussions in Section \ref{sec:ethical} regarding the use of AI4EO for social good, which involves applying AI technologies to address and solve social, economic and environmental challenges such as healthcare, education, environmental sustainability, poverty alleviation and disaster response. By leveraging AI technologies and the ethical considerations mentioned in the previous sections, these initiatives seek to improve outcomes and make a positive difference in people's lives, which is the ultimate goal of responsible AI.

Although the applications of AI4EO for social good are extensive \citep{persello2022deep,burke2021using}, we focus here primarily on understanding disaster risks and evaluating their impacts through several key examples crucial for effective disaster management and financing strategies \cite{SocialGood_IPCC22,SocialGood_IPCC21,SocialGood_AdaptPathways}. Robust frameworks are necessary for investing in projects that enhance resilience to climate change and its associated disasters. AI, leveraging integrated EO data, can significantly contribute to building these frameworks by analyzing such data through innovative computational methods \cite{SocialGood_AI,SocialGood_IPCC22,SocialGood_IPCC21}. 
However, it is also crucial to implement policies and decisions at the local level, as the impacts of climate change vary over time and across different regions \cite{SocialGood_IPCC22,SocialGood_ECMWF,SocialGood_GreenDeal,SocialGood_SG_GreenPlan,SocialGood_USlegisl}.

Understanding disaster risks and evaluating their impacts are crucial for effective disaster management and financing strategies \cite{SocialGood_IPCC22,SocialGood_IPCC21,SocialGood_AdaptPathways}. Robust frameworks are necessary for investing in projects that enhance resilience to climate change and its associated disasters. AI leveraging integrated EO data can contribute significantly to building these frameworks by analyzing such data through innovative computational methods \cite{SocialGood_AI,SocialGood_IPCC22,SocialGood_IPCC21}. 
 Moreover, it is also crucial to address the implementation of policies and decisions at the local level, as the impacts of climate change vary over time and across space \cite{SocialGood_IPCC22,SocialGood_ECMWF,SocialGood_GreenDeal,SocialGood_SG_GreenPlan,SocialGood_USlegisl}. 
For example, rising temperatures in Europe can intensify evapotranspiration, potentially causing more floods in Germany and water shortages in Italy. Meanwhile, Portugal, France, and Greece may face increased wildfires due to similar processes 
\cite{SocialGood_ECMWF,SocialGood_ESA,SocialGood_IPCC22,SocialGood_GreenDeal}. Global warming also impacts the water cycle, necessitating increased irrigation for vegetation health and food security. This can heighten the occurrence of natural hazards, even across significant distances. For example, irrigation in North America's Central Valley, High Plains, and Mississippi River Valley affects precipitation in regions like the Colorado River, USA Midwest, and Southeast, respectively \cite{SocialGood_Irrig1,SocialGood_Irrig2,SocialGood_Irrig3,SocialGood_Irrig4}.

The above instances emphasize the diverse impacts of climate change on different regions, even at relatively close range. Thus, it is crucial to enable each region to be prepared and ready to implement specific contingency plans based on local changes to environmental events characterized by changes in atmospheric and oceanographic variables \cite{SocialGood_AdaptPathways,SocialGood_ECMWF,SocialGood_GreenDeal,SocialGood_IPCC22}. 
It is also worth noting that these changes might affect the socio-economic fabric of diverse biogeographic regions, thereby influencing the stability and security of human welfare and demographic fluxes \citep{SocialGood_AdaptPathways,SocialGood_ECMWF,SocialGood_GreenDeal,SocialGood_IPCC21,SocialGood_IPCC22,SocialGood_SG_GreenPlan}. 
It is, therefore, paramount for AI to provide a deep understanding of EO data so that decision- and policymakers are enabled to implement climate change adaptation and mitigation strategies. These strategies affect investments and solutions that citizens, local authorities, and businesses can undertake. Additionally, it implies the ability to engage stakeholders and public opinion to gain societal consensus 
\cite{SocialGood_ECMWF,SocialGood_ESA,SocialGood_AI,SocialGood_GreenDeal}.

To provide concrete examples of the actual impact and possibilities unlocked by the appropriate use of AI in decision-making processes, we present two instances of operational scenarios that would benefit greatly from a deep understanding of Earth processes to enhance human welfare and improve communities' wellbeing.

\subsection{Early Warning Systems for Mass Movements}

One of the most severe effects of climate change is the increased occurrence of catastrophic events caused by mass movements. This affects human welfare and results in huge economic losses. As such, the development of early warning systems (EWSs) is crucial to the everyday decisions of local governments and communities, and exemplify Responsible AI by prioritizing human safety and equity through timely alerts and accessible technology \cite{SocialGood_AdaptPathways,SocialGood_ECMWF,SocialGood_GreenDeal,SocialGood_IPCC21,SocialGood_IPCC22}. 

Historically, the main objective of EWSs has been to quantify the degree of immediate danger that every human settlement in each region of interest is exposed to (see, \cite{azarafza2021deep,SocialGood_EWS1,SocialGood_EWS2,SocialGood_EWS3}). In fact, state-of-the-art EWSs used in operational pipelines are based on the analysis of data and measurements at the regional level, which are then aggregated to avoid issues related to the irregularity of the temporal and spatial scale at which each record can be acquired. In this way, state-of-the-art EWSs are prone to highly sensitive and coarse classifications and predictions, potentially leading to a poor perception of risk and inadequate local management plans. Moreover, the classic structure of operational EWSs does not account for cascading effects affecting local communities at social and economic levels. For example, state-of-the-art EWSs for mass movements do not consider a town that can be cut off from transportation and communication systems because of landslides blocking the roads around it as relevant for early warning \cite{SocialGood_EWS3}. 

AI4EO can be a game-changer in this arena \cite{SocialGood_ESA,SocialGood_ECMWF}. Recent studies \cite{SocialGood_EWS_AGU23} have shown the improvement to state-of-the-art EWSs' capacity provided by analysis of the environmental (hydrological and geological) characteristics of mass movements derived from remote sensing platforms and the connectivity information of formal settlements and road network data in terms of graph structures. This architecture allows us to study the interaction of the probabilistic mass movement susceptibility (derived from the environmental properties by means of a supervised ensemble graph neural network) on the graph representing the communication network (that could involve roads, electricity grids, water pipelines) connecting the formal settlements. Investigating this structure by graph-based methods (e.g., spectral clustering) results in the derivation of indices to quantify the probability of each human settlement to be directly or indirectly affected by mass movements (i.e., the probability to be hit by a mass movement event - such as a landslide - or the probability to be isolated as a result of mass movements affecting their surroundings) \cite{SocialGood_EWS_AGU23}. This approach has the capacity to pave the way for measures and policies for adaptation and mitigation through a new holistic graphical perspective to assess various large-scale geospatial datasets of risk elements such as exposure, vulnerability and hazard. In particular, it has been shown that this strategy can deliver robust and reliable outcomes for hindcasting and forecasting of the impact of mass movements at different scales (local, regional, and national), able to explore long temporal trends (e.g., over 68,000 incidents of reported mass movements since 1957) and accurately identify the critical factors for climate change-induced mass movements \cite{SocialGood_EWS_AGU23}.

\begin{figure}[htb]
	\centering
	\includegraphics[width=.5\columnwidth]{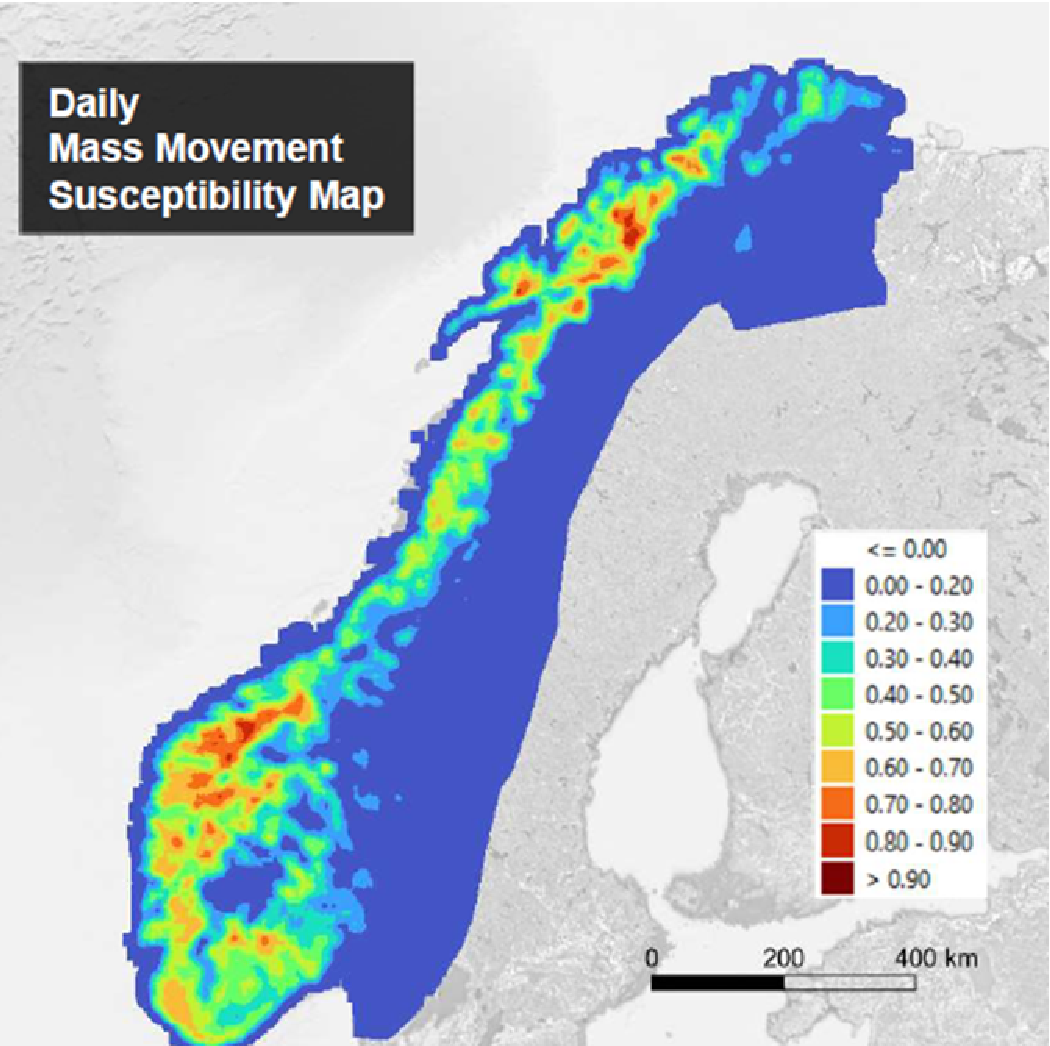}
	\caption{Estimation of mass movements susceptibility index across the Norwegian territory by means of the architecture proposed in \cite{SocialGood_EWS_AGU23}.}
	\label{SocialGood_EWS}
\end{figure}

\subsection{Climate Teleconnections}

One of the most important open challenges in environmental monitoring and climate change studies is represented by the analysis of climate teleconnections, that is, the investigation of the interplay between weather phenomena at widely separated locations on Earth. This implies the study of climate patterns that span thousands of kilometers. Examples of teleconnection phenomena include shifts of atmospheric mass and/or pressure back and forth between two distant locations, where changes in atmospheric pressure in one location lead to corresponding changes in pressure in another location far away \cite{SocialGood_Telecon7,SocialGood_Telecon8,SocialGood_Telecon9}. Therefore, responsible AI-driven insights are crucial for informing global climate strategies and policies to ensure that all communities benefit equitably.

The water cycle is one of the primary drivers of climate teleconnections worldwide. In this respect, the three world poles (i.e., Arctic, Antarctic, high mountain Asia (HMA)) play a key role in influencing global weather patterns through their impact on teleconnections \cite{SocialGood_Telecon1,SocialGood_Telecon2,SocialGood_Telecon3,SocialGood_Telecon4}. Specifically, the weather and climate of HMA and the surrounding region have been demonstrated to be influenced by global modes of variability (MoV), which are fluctuations of atmospheric and oceanographic variables (e.g., sea surface temperature, rainfall, surface pressure, wind speed) caused by the continuous exchange of mass, momentum and energy on all timescales between Earth’s atmosphere, oceans, cryosphere and continental hydrology. For example, the planetary-scale Rossby-wave propagation that causes the intra-seasonal variability over central Asia and the northern part of India has been demonstrated to be affected by the Eurasian teleconnection. Also well-known is the interplay and impact on HMA climate variations of the Indian Ocean Dipole (IOD) and El Niño Southern Oscillation (Nino34), North Atlantic Oscillation (NAO), the Central Indian Ocean mode, and the boreal summer intraseasonal oscillation. However, a thorough understanding of the historical behavior of global MoV and its influence on weather patterns within the HMA has not yet been derived \cite{SocialGood_Telecon1,SocialGood_Telecon2,SocialGood_Telecon3,SocialGood_Telecon4,SocialGood_Telecon5,SocialGood_Telecon6,SocialGood_Telecon8,SocialGood_Telecon7}.

The use of advanced technologies for machine learning, relying on records acquired by EO platforms, plays a key role \cite{SocialGood_Telecon9,SocialGood_Telecon7}. In fact, as remote sensors can acquire properties of the previously mentioned environmental variables with precision at a global scale, they enable the investigation of MoVs that drive climate teleconnections worldwide. In particular, when considering HMA, the integration of remote sensing data with hydrometeorological climate variables (e.g., geopotential height at 250 hPa (z250), 2-m air temperature (T2M), total precipitation (PRECTOT), and fractional snow cover area (fSCA)) leads to a comprehensive exploration of the interactions across several MoVs over HMA (e.g., the Eurasian teleconnection, the IOD, the NAO and the Nino34). This results in the quantification of correlations and cause/effect scenarios across MoVs and climate variables, as well as the predictability of weather and climate patterns of the variables of interest, so that proper planning of resources and policies to address and adapt to climate change can be put in place \cite{SocialGood_ESA,SocialGood_AdaptPathways,SocialGood_GreenDeal,SocialGood_IPCC22,SocialGood_IPCC21,SocialGood_SG_GreenPlan,SocialGood_ECMWF}. 

	\section{Responsible AI Integration in Business Innovation and Sustainability}
	\label{sec: business}
	
	The integration of AI with low-cost, high-performance computing is fostering business innovation, with organizations spanning various sectors and industries, disrupting both commercial and sustainability-focused business models. These may include monitoring remote infrastructure, assessing climate risk, dynamically managing supply chains and decision-making processes across various industries. For example, an oil and gas company could enhance the monitoring of its infrastructures, such as the pipelines used for the transportation of natural gas, crude and refined petroleum by incorporating geotagged data into the operational workflow of the status and health of its distributed systems. These insights could facilitate timely and cost-effective decisions, guiding preventive maintenance efforts to reduce oil spills and gas leakages.
	However, the use of AI-based solutions requires appropriate governance and safeguards to mitigate potential negative impacts. Therefore, emphasizing governance, standards, open-source solutions, and innovative business models is crucial to foster the fair adoption of AI for EO. For example, while satellite data proves invaluable in agricultural management, farmers may lack direct access to the collected data. In contrast, business competitors with greater resources may access this information and potentially use it to acquire land at favorable terms. On the other hand, biases embedded within AI models may lead to outcomes that lack equity and inclusivity across different regions and populations. For example, the use of farm imagery without agricultural knowledge could result in the implementation of costly regulations or insurance premiums, adversely affecting farmers.
	The integration of AI-based solutions into climate action and sustainable growth provides consistent, objective measurements, fostering trust and transparency in efforts toward achieving a net-zero economy. The convergence of AI and Environmental, Social, and Governance (ESG) factors represents a significant milestone in today's dynamic business environment. In this context, sustainability refers to how a company's business model, including its products and services, contributes to sustainable development. By incorporating AI into ESG risk management and strategies, commercial data providers can achieve their sustainability objectives and explore novel pathways for fostering sustainable development and innovation. ESG reports are generally published yearly by commercial satellite providers such as Maxar Technologies \cite{maxarMaxarPublishes}, Planet \cite{planet} and Airbus \cite{airbusSustainability}, and technology companies such as Microsoft \cite{microsoftGoodMicrosoft}, Google \cite{sustainabilitySustainableInnovation}, and IBM \cite{ibmPlanetPeople}. Some examples of ESG-oriented use-cases of AI in Earth observation include:
	\begin{itemize}
		\item	\textbf{Carbon footprint of AI models:} the growing carbon footprint of the AI industry inspired the University of Copenhagen to develop “carbontracker” \cite{anthony2020carbontracker}, a tool that estimates the carbon emissions associated with different hardware and software configurations, and forecasts training runs' carbon footprint. It also supports predicting the total duration, energy, and carbon footprint of training an AI model. Then, the carbon intensity is used to indicate the carbon footprint in ESG reports.
		\item	\textbf{Distribution of open-source training data:} SpaceNet \cite{spacenetSpacenetaix2013} is a nonprofit consortium dedicated to accelerating the research and application of open-source AI technology for geospatial applications. Co-founded by CosmiQ Works and Maxar Technologies, SpaceNet includes partners such as Amazon Web Services (AWS), Capella Space, IEEE Geoscience and Remote Sensing Society (GRSS), National Geospatial-Intelligence Agency (NGA), Oak Ridge National Lab, Open Geospatial Consortium (OGC), Planet, Topcoder, and Umbra. SpaceNet offers open, precision-labeled, electro-optical and synthetic aperture radar satellite datasets and runs challenges to encourage the development of algorithms designed specifically for geospatial applications.
		\item	\textbf{Integrating diverse datasets to facilitate comprehensive situation and monitoring changes over time:} Microsoft’s Planetary Computer \cite{microsoftMicrosoftPlanetary} combines a multi-petabyte catalog of global environmental data that be queried via APIs, a scientific environment that allows users to answer global questions about that data, and applications that put those answers in the hands of conservation stakeholders.
		\item	\textbf{Identifying sources of environmental issues, such as methane emissions from landfills, farms, or pipelines:} Carbon Mapper \cite{planetCarbonMapper} is a nonprofit organization co-founded by the State of California, NASA’s Jet Propulsion Laboratory (JPL), Planet, the University of Arizona, Arizona State University (ASU), High Tide Foundation, and RMI that plans to deploy a hyperspectral satellite constellation with the ability to pinpoint, quantify and track point-source methane and CO2 emissions.
	\end{itemize}

	\section{Conclusions, Remarks and Future Directions}
	\label{sec: conclusion}
	
	\subsection{Conclusions and Remarks}
	
	This paper addresses the pressing need to navigate the ethical and societal complexities inherent in AI4EO applications. By synthesizing insights from diverse sources, including the academic literature, business models and ethical frameworks, it provides actionable guidance for researchers and practitioners in this rapidly evolving field. The paper's significance lies in its comprehensive examination of ethical dimensions such as privacy, security, fairness, responsibility and mitigating (unfair) biases, shedding light on key considerations that underpin responsible AI practices in EO missions. Through fostering collaboration and dialogue among stakeholders, the paper aims to advance ethical discourse and promote a culture of responsible AI development and deployment. It serves as a foundational framework for navigating ethical dilemmas and guiding the responsible integration of AI technologies into EO missions. Ultimately, the paper underscores the transformative potential of AI4EO applications in addressing global challenges while emphasizing the importance of upholding ethical principles. However, it is crucial to acknowledge that in the current context of significant global conflicts, some state actors may resort to non-peaceful means to achieve their own interests over the collective good defined at the global level. Ensuring that these technologies contribute to a more equitable and sustainable world requires navigating these complex realities.
	
	Given the current geopolitical state of the world, the role of AI in Earth observation (AI4EO) takes on heightened significance in contributing to global stability and peace. \textbf{Aligning with UNSDG 16, which emphasizes peace, justice, and strong institutions, this work underscores the critical importance of leveraging AI4EO to support these goals and mitigate conflict.}
	
	Furthermore, the research paper explores the significant effects of AI4EO applications on political, societal and environmental landscapes. It frames the United Nations Sustainable Development Goals as the guiding North Star for the ethical use of AI4EO, emphasizing that these goals should define what 'good' looks like. By aligning AI4EO efforts with the SDGs, the paper sets a clear benchmark for transparency, reproducibility and trust in AI4EO research. \textbf{The paper advocates for open access to data and code to facilitate evidence-based decision-making processes in policy-making, enabling stakeholders to make well-informed choices.} Societally, it highlights the \textbf{critical need to address biases and safeguard privacy rights to promote fair outcomes and reduce stigmatization in vulnerable communities}. The emphasis on responsible AI practices prioritizes fairness and inclusivity, fostering societal trust and cohesion. Environmentally, the paper stresses \textbf{AI's vital role in addressing pressing environmental challenges through sustainable resource management and conservation efforts}. By aligning AI4EO workflows with sustainability principles, the paper offers a path for leveraging AI technologies to advance the SDGs, thereby contributing to a more resilient and sustainable future for humanity and the planet.
	
	\subsection{Future Investigations}
	
	Future investigations, as indicated multiple times throughout this paper, should focus on developing \textbf{strategies that facilitate the open sharing of data} for scientific advancements while simultaneously \textbf{preserving sensitive personal or group information}. Ensuring the ethical use and protection of such data is crucial, particularly in applications like mapping deprived urban areas or victim detection, where the responsible use of AI can have significant societal impacts. These applications critically depend on the availability of large benchmark datasets, which are currently lacking. To promote responsible AI, it is essential to \textbf{implement robust privacy-preserving techniques and data anonymization methods} that protect individuals' identities while maintaining the utility of the data for research purposes. This includes using techniques like differential privacy, federated learning, and secure multiparty computation to safeguard sensitive information. Additionally, it is vital to establish clear guidelines and ethical standards for data sharing and usage. These guidelines should ensure that data are used in a manner that respects the rights and dignity of individuals and communities. \textbf{Collaboration with stakeholders, including ethicists, legal experts and representatives from affected communities, is necessary} to create frameworks that balance the need for scientific progress with the imperative to protect sensitive information. Moreover, \textbf{transparency in data collection, processing and sharing practices is essential}. Researchers and organizations must be open about the sources of their data, the methods used to process it, and the steps taken to protect privacy. This transparency builds trust and ensures that AI advancements are grounded in ethical practices. By prioritizing the responsible handling of sensitive data, the research community can develop AI-based methods that contribute to societal well-being while respecting individual and group privacy. 
	
	\textbf{In the early 2020s, following decades of model-centric approaches and the deep learning revolution, there was a notable shift from model-centric to data-centric learning.} The concept of data-centric AI began gaining traction, emphasizing the quality of data over the complexity of models. Concurrently, a strong societal push emerged with the advent of Responsible AI recommendations and legislation, as outlined in this paper. This necessitates research to develop methods that can capture biases in compliance with such legislation while maintaining scientific robustness and addressing the unique characteristics of GRS data. \textbf{Responsible AI must consider the entire machine learning pipeline and involve multiple stakeholders.} This includes identifying sensitive attributes and integrating specific domain knowledge into bias mitigation techniques. \textbf{Currently, there is a balanced focus on both data quality and model innovation (the era of data-model-centric approaches).} Techniques such as foundation models, synthetic data generation, uncertainty modeling and explainability, and robust data validation are employed to enhance data-centric approaches, emphasizing both model and data equally. Ethical considerations, including fairness, transparency, and bias mitigation, are also integral to these AI practices.
	

	
	\printbibliography
	\clearpage
	
\end{document}